\documentclass{article}
\usepackage{spconf,amsmath,epsfig}
\usepackage{multirow}
\makeatletter
\newcommand{\mypm}{\mathbin{\mathpalette\@mypm\relax}}
\newcommand{\@mypm}[2]{\ooalign{%
  \raisebox{.1\height}{$#1+$}\cr
  \smash{\raisebox{-.6\height}{$#1-$}}\cr}}
\makeatother
\usepackage{bbold}
\usepackage{float}
\restylefloat{table}

\title{Segmentation-Aware Hyperspectral Image Classification}
\name{Berkan Demirel, Omer Ozdil, Yunus Emre Esin, Safak Ozturk}
\address{Image and Video Processing Group, HAVELSAN Inc., Ankara, TURKEY \\
\{bdemirel, oozdil, yesin,  sozturk\}@havelsan.com.tr}

\begin{document}
%
\maketitle
\begin{abstract}
In this paper, we propose an unified hyperspectral image classification method which takes three-dimensional hyperspectral data cube as an input and produces a classification map. In the proposed method, a deep neural network which uses spectral and spatial information together with residual connections, and pixel affinity network based segmentation-aware superpixels are used together. In the architecture, segmentation-aware superpixels run on the initial classification map of deep residual network, and apply majority voting on obtained results. Experimental results show that our propoped method yields state-of-the-art results in two benchmark datasets. Moreover, we also show that the segmentation-aware superpixels have great contribution to the success of hyperspectral image classification methods in cases where training data is insufficient.
\end{abstract}
\section{Introduction}
\label{sec:intro}

Hyperspectral images are obtained by dividing the electromagnetic spectrum into several narrow bands that can be extended beyong the visible light. Thanks to this feature, strong spectral classification of surfaces and objects are possible through hyperspectral data. Hyperspectral imaging technology is currently used in many areas such as  target detection~\cite{manolakis2003hyperspectral}, plant species detection~\cite{ozdil2018representative,demirel2017vegetation}, and water property detection~\cite{lee1999hyperspectral}. In these studies, various machine learning techniques have been used. More recently, convolutional neural network (ConvNet) based approaches have lead to great advances in some hyperspectral imaging problems, especially in target classification~\cite{zhong2018spectral, xu2018spectral}.

Deep learning based approaches require large amount of labeled images (\textit{e.g.} ImageNet~\cite{deng2009imagenet}) in the classification problem. In the field of hyperspectral imaging, it is not possible to obtain large amount of images due to sensor and preprocessing costs. Moreover, deep neural network approaches  for hyperspectral classification problem use certain amount of spectral signatures collected from same image with test set for training purpose and it leads to biased deep classification models. Therefore, it is necessary to focus on transfer learning approaches in hyperspectral imaging area.

It is difficult to transfer information in the spectral sense, but it is possible to transfer spatial information which are useful for encoding context information from three-band datasets~\cite{demirel2016hyperspectral}. In this paper, we try to transfer spatial information by using pixel affinity network based segmentation-aware superpixels~\cite{liu2018learning}. Pixels that are spatially close each other and have similar texture or color information will belong to the same class and according to this hypothesis, we can assume that pixels that are located in a segmentation-aware superpixel belong to the same class. Therefore, we can correct the erroneous class predictions obtained from deep neural networks by conducting dominance analysis on the superpixel groups. In this model, pixel affinity network is trained on some images of BSDS500 dataset~\cite{arbelaez2011contour}. As a deep convolutional model, we use an architecture which is identical to that described in~\cite{zhong2018spectral}. The proposed deep model learns spectral and spatial representations separately by using two sequential residual blocks. We evaluate the proposed method on University of Pavia and Indian Pines datasets and experimental results show that our propoped method yields state-of-the-art results in these two benchmark datasets.

\begin{figure*}
  \includegraphics[width=\textwidth,height=4cm]{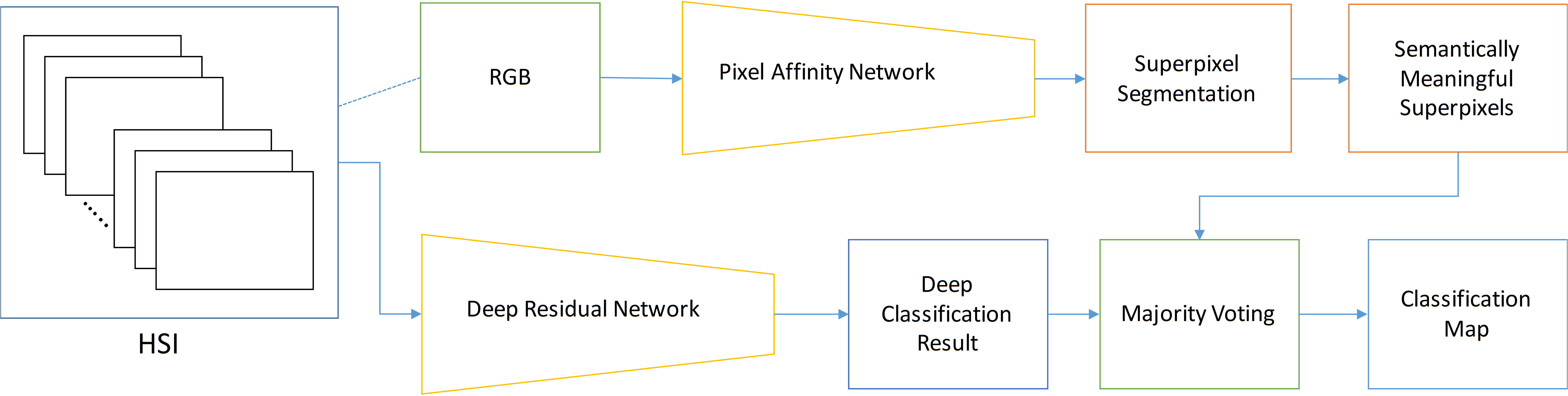}
  \caption{Our classification framework. It uses unified classification method which consists of a deep neural network that uses spectral and spatial information together with residual connections and pixel affinity network based on segmentation-aware superpixels together.}
   \label{main_figure}
\end{figure*}

To sum up, we proposed an unified classification method that consists of a deep neural network which uses spectral and spatial information together with residual connections and pixel affinity network based on segmentation-aware superpixels together. Our unified framework is summarized on Figure~\ref{main_figure}. Our main contributions related to the classification of hyperspectral images as follows: 
\begin{enumerate}
  \item We transfer the spatial information from three-band datasets (\textit{e.g.} BSDS500) for encoding context information.
  \item We learn with limited spectral training data by using this spatial information and compare our model with state-of-the-art models in the same setting.
  \item The proposed method obtains the state-of-the-art performance among hyperspectral image classification algorithm for two benchmark datasets.
\end{enumerate}

In the rest of the paper, we first describe the proposed method in Section~\ref{sec:method} and then talk about experimental setup, and compare experimental results in Section~\ref{sec:experiments}. Finally, we will summarize the motivation behind this paper and discuss other things in Section~\ref{sec:conclusion}.

\section{Method}
\label{sec:method}
In this section we describe the proposed method. Our classification model needs three-dimensional hyperspectral data cube and RGB version of it to generate a classification map in two subsections: (1) RGB image is used as an input for segmentation-aware superpixels, and (2) hyperspectral data is used as an input in deep neural network. In order to model this process, we first train a deep neural network which uses spectral and spatial information together with residual connections. 

The neural network predicts a classification result of hyperspectral data. This process is carried out in parallel with the generation of the segmentation-aware superpixels over the RGB image. All the spectral signatures within the superpixel will have the same color and texture information and therefore belong to the same class. We use them to compensate for the incorrect classification results obtained from the deep neural network. For this purpose, we perform a dominance analysis within each superpixel and transfer the dominant class information to all pixels within the superpixel. Computing superpixels with learned pixel affinities provides an opportunity to preserve object boundaries.

Our approach can be defined as follows: let $X_{H}$ be the hyperspectral data cube, $X_{RGB}$ be the RGB version of $X_{H}$, $Z$ be the classification map obtained from deep neural network and $Y$ be the final classification map. We define a function $f(x, W):X_{H} \rightarrow R^{m \times n \times \lambda}$ for classification results that are obtained from deep neural network:

\begin{equation}
Z = f(X_{H},W)
\end{equation}

where $W$ is weights for trained deep network which uses spectral and spatial information together with residual connections. The neural network used in this paper is identical to the model described in~\cite{zhong2018spectral}. 

We define an another function $g(x, n):X_{RGB} \rightarrow R^{m\times n}$ for generating segmentation-aware superpixels:

\begin{equation}
S = g(X_{RGB}, n)
\end{equation}

where $n$ is the number of expected superpixels. Here, $g$ is a pixel affinity network based segmentation-aware superpixel algorithm. Pixel affinity network is trained on BSDS500 dataset~\cite{arbelaez2011contour} for preserving object boundaries by using segmentation-aware loss in superpixel generation. The details of the pixel affinity network and segmentation-aware superpixels are described in~\cite{liu2018learning}. 

We compute the dominant class information for each superpixel and apply this class information all pixels within the related superpixel:

\begin{equation}
Y_{i,j} = \phi(Z_{i,j},S_{k})
\end{equation}

where $Y_{i,j}$ is the final classification result of pixel location $(i,j)$, and $Z$ is the classification result of the deep neural network. Moreover, $S_{k}$ represent the superpixel group that consist of $i,j$ coordinate. Here, $\phi(Z_{i,j}, S_{k})$ is defined as follows:

\begin{equation}
\phi(Z_{i,j},S_{k}) = \underset{t}{\mathrm{argmax}}\frac{\sum_{t=1}^{C}\sum_{p=1}^{m \times n} \left [ \mathbb{1}_{p}^{S_{k}}\times \mathbb{1}_{Z_{p}}^{C_{t}}\right ]}{\sum_{p=1}^{m \times n} \mid  \mathbb{1}_{p}^{S_{k}}\mid }
\end{equation}

where $C_{t}$ represents the candidate class, $p$ represents the current pixel and $\mathbb{1}_{p}^{S_{k}}$ notation represents the scalar value that 1 if the selected pixel is located in $S_{k}$ superpixel group, otherwise 0. Moreover, $\mathbb{1}_{Z_{p}}^{C_{t}}$ represents the value that 1 if the classification result of pixel $p$ equals to the label of the class $C_{t}$, otherwise 0.

\begin{table*}[]
\begin{center}
\label{comparison}
 \caption{Overal accuracies of University of Pavia and Indian Pines datasets.}
\begin{tabular}{|l|l|l|l|l|l|l|l|l|l|}
\hline
                                     & Raw   & SVM\cite{waske2010sensitivity}   & SAE\cite{chen2014deep}  & EMP\cite{benediktsson2005classification}   & LeNet\cite{lecun1998gradient} & CNN\cite{chen2016deep}& SSUN\cite{xu2018spectral}  & SSRN\cite{zhong2018spectral}  & Our Method \\ \hline
\multirow{2}{*}{Uni. of Pavia} & 90.80 & 90.58 & 94.25 & 97.61 & 97.73 & 98.64 & 99.46 & 99.79 &\textbf{99.93}  \\ \cline{2-10} 
                                     & $\mypm$ 0.37 & $\mypm$0.47 & $\mypm$0.18 & $\mypm$0.19 & $\mypm$0.96 & $\mypm$0.20 & $\mypm$0.32 & $\mypm$0.09 & $\mypm$\textbf{0.02}   \\ \hline
\multirow{2}{*}{Indian Pines}        & 79.68 & 81.67 & 85.47 & 92.22 & 96.01 & 97.41 & 98.40 & 99.19 & \textbf{99.38}  \\ \cline{2-10} 
                                     &$\mypm$ 0.87 & $\mypm$0.65 & $\mypm$0.58 & $\mypm$0.71 & $\mypm$0.87 & $\mypm$0.43 & $\mypm$0.37 & $\mypm$0.26 & $\mypm$\textbf{0.06}   \\ \hline
\end{tabular}
\end{center}
\end{table*}

\section{Experiments}
\label{sec:experiments}

In this section, we present the effectiveness of the proposed approach by considering two different hyperspectral image classification datasets: University of Pavia and Indian Pines. 

In the rest of the section, we first describe the datasets in Section~\ref{sec:experiments1}. Then, we give the implementation details and experimental results in Section~\ref{sec:experiments2}. Finally, we will discuss ablative studies on training data in Section~\ref{sec:experiments3}.

\subsection{Datasets}
\label{sec:experiments1}
\textbf{University of Pavia} dataset has $610\times340$ pixels with 9 different land-cover class types. It is collected from the reflective optics system  imaging spectrometer (ROSIS-03) and its spatial resolution is 1.3m by pixels. In our experiments, we use $0.5\%$, $5\%$ and $10\%$ of total pixels for training purpose separately.

\textbf{Indian Pines} dataset has $145\times145$ pixels with 16 different vegetation class types. It is acquired by the Airborne Visible/Infrared Imaging Spectrometer (AVIRIS) and its spatial resolution is 20m by pixels. For the experiments of the Indian Pines dataset, we use $0.5\%$, $5\%$ and $20\%$ of number of total pixels for training purpose separately.

\subsection{Experimental Results}
\label{sec:experiments2}
In our experiments, we select number of superpixels as $n=10.000$ to obtain fine-grained superpixel groups. We first compare the performance of the proposed approach with other state-of-the-art results by running the proposed method with same percentage of training data. In these experiments, training of both University of Pavia and Indian Pines models are completed in $40$ epoches. Moreover, we select batch size as 16 and learning rate as 0.001.

We compare our proposed approach with SVM\cite{waske2010sensitivity}, SAE\cite{chen2014deep}, EMP\cite{benediktsson2005classification} , LeNet\cite{lecun1998gradient}, CNN\cite{chen2016deep}, SSUN\cite{xu2018spectral}  and SSRN\cite{zhong2018spectral} methods. The overall accuracy results are given in Table~\ref{comparison}. In these experiments, we use $10\%$ of total pixels in University of Pavia and $20\%$ of total pixels in Indian Pines datasets for training purpose. We observe that our method obtain state-of-the-art results on both datasets. 

The results on Table~\ref{comparison} show the importance of the context information in hyperspectral image classification problem. Moreover, these results also show that we can provide context information with segmentation-aware superpixels which are trained on three-band computer vision datasets. We share our final classification map for University of Pavia dataset on Figure 2.

\begin{figure}[]

\begin{minipage}[b]{0.28\linewidth}
  \centering
\centerline{\epsfig{figure=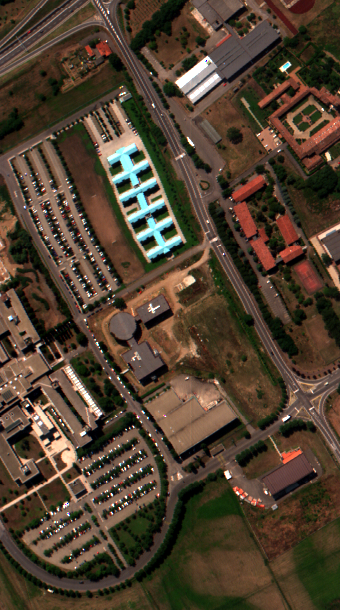,width=2.7cm}}
  \vspace{.2cm}
  \centerline{(a)}\medskip
\end{minipage}
\hfill
\begin{minipage}[b]{.28\linewidth}
  \centering
 \centerline{\epsfig{figure=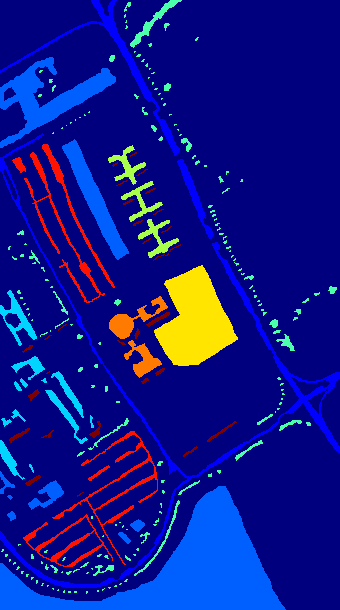,width=2.7cm}}
  \vspace{0.2cm}
  \centerline{(b)}\medskip
\end{minipage}
\hfill
\begin{minipage}[b]{0.28\linewidth}
  \centering
 \centerline{\epsfig{figure=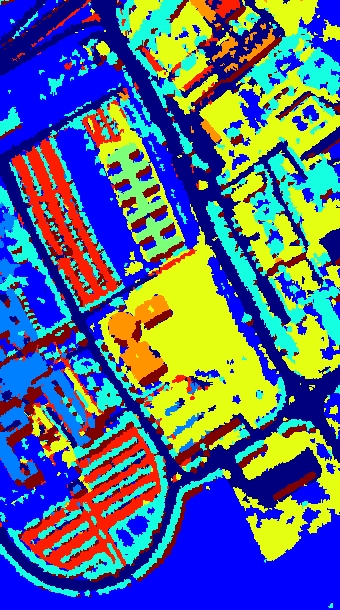,width=2.7cm}}
  \vspace{0.2cm}
  \centerline{(c)}\medskip
\end{minipage}
\caption{Our visual results on University of Pavia dataset. (a) represents original image, (b) is the groundtrurth map, and (c) is our final classification map when we use $10\%$ of total pixels for training purpose.}
\end{figure}

\subsection{Ablative Studies on Training Data}
\label{sec:experiments3}
In this section, we will discuss ablative studies on training data. One of the contributions we have made in this paper is we show that spatial information which are obtained from segmentation-aware superpixels are very useful in situations where there is a limited training set for hyperspectral image classification. 

We compared our method with the SSRN~\cite{zhong2018spectral} by using various amounts of training set. We use University of Pavia and Indian Pines datasets during the experiments and we share our results on Table 2 and Table 3. From these results, we see that segmentation-aware superpixels contribute to the classification models that have a certain level of success.

\begin{table}[]
\begin{center}
 \caption{Overal accuracies of University of Pavia dataset for different percentage of training data.}
\begin{tabular}{|l|l|l|l|}
\hline
                            & 0.5\% & 5\%   & 10\%  \\ \hline
\multirow{2}{*}{SSRN}       & 90.01 & 99.24 & 99.79 \\ \cline{2-4} 
                            & $\mypm$0.02 & $\mypm$0.03 & $\mypm$0.09 \\ \hline
\multirow{2}{*}{Our Method} & 90.47 & 99.36 & 99.93 \\ \cline{2-4} 
                            & $\mypm$0.04 & $\mypm$0.06 & $\mypm$0.02 \\ \hline
\end{tabular}
\end{center}
\end{table}

\begin{table}[]
\begin{center}
 \caption{Overal accuracies of Indian Pines dataset for different percentage of training data.}
\begin{tabular}{|l|l|l|l|}
\hline
                            & 0.5\% & 5\%   & 20\%  \\ \hline
\multirow{2}{*}{SSRN}       & 30.52 & 91.99 & 99.19 \\ \cline{2-4} 
                            & $\mypm$0.13 & $\mypm$0.14 &$\mypm$0.26 \\ \hline
\multirow{2}{*}{Our Method} & 31.54 & 92.14 & 99.38 \\ \cline{2-4} 
                            & $\mypm$0.20 & $\mypm$0.12 &$\mypm$0.06 \\ \hline
\end{tabular}
\end{center}
\end{table}

\section{Conclusion}
\label{sec:conclusion}
In this paper, we proposed a hyperspectral image classsification method which use pixel affinity network based segmentation-aware superpixels and deep neural network in a unified framework. Several experimental results show that the proposed method yields state-of-the-art results in two benchmark datasets. Also, ablative studies show that the segmentation-aware superpixels have great contribution to the success of classification in cases where training data are insufficient.

\bibliographystyle{IEEEbib}
\bibliography{paper}

\end{document}